# A Methodology for Dynamic Parameters Identification of 3-DOF Parallel Robots in Terms of Relevant Parameters

Miguel Díaz-Rodríguez[1], Vicente Mata[2], Ángel Valera[3], Álvaro Page[4]

[1]*Departamento de Tecnología y Diseño, Facultad de Ingeniería, Universidad de los Andes, 5101a, Núcleo La Hechicera, Mérida, Venezuela, dmiguel@ula.ve*
[2]*Departamento de Mecánica y Materiales, Universidad Politécnica de Valencia, 46022, Camino de Vera, Valencia, Spain, vmata@mcm.upv.es*
[3]*Departamento de Ingeniería de Sistemas y Automática, Universidad Politécnica de Valencia, 46022, Camino de Vera, Valencia, Spain, giuprog@isa.upv.es*
[4]*Departamento de Física Aplicada, Universidad Politécnica de Valencia, 46022, Camino de Vera, Valencia, Spain, alvaro.page@ibv.upv.es*

**Abstract**

The identification of dynamic parameters in mechanical systems is important for improving model-based control as well as for performing realistic dynamic simulations. Generally, when identification techniques are applied only a subset of so-called base parameters can be identified. More even, some of these parameters cannot be identified properly given that they have a small contribution to the robot dynamics and hence in the presence of noise in measurements and discrepancy in modeling, their quality of being identifiable decreases. For this reason, a strategy for dynamic parameter identification of fully parallel robots in terms of a subset called relevant parameters is put forward. The objective of the proposed methodology is to start from a full dynamic model, then simplification concerning the geometry of each link and, the symmetry due to legs of fully parallel robots, are carried out. After that, the identification is done by Weighted Least Squares. Then, with statistical considerations the model is reduced until the physical feasibility conditions are met. The application of the propose strategy has been experimentally tested on two different configurations of actual 3-DOF parallel robots. The response of the inverse and forward dynamics of the identified models agrees with experiments. In order to evaluate the forward dynamics response, an approach for obtaining the forward dynamics in terms of the relevant parameters is also proposed.

---

## 1. Introduction

Parallel robots have constituted a very active field of research over the last 20 years. Compared to serial robots, parallel robots have essentially two well-known advantages, namely greater precision in positioning and increased rigidity with respect to the relationship between size and workload limit. The main drawbacks of parallel robots are their small workspace and some specific problems related to control planning. Although their implementation has being focused on academia, nowadays their application is being transferring to industry [1]. Thus, researches for improving and developing accurate dynamics models for these class of robots need to be carried out, particularly for those with less than 6-DOF.

The relevance of dynamic parameter identification lies in the fact that accurate identification of the dynamic parameters underlying the dynamic model is necessary in order to obtain realistic dynamic simulation of mechanical systems. On the other hand, the identified parameters can be used for the development of advanced model-based control schemes. The dynamic parameters of the robot to be identified are basically mass, location of the center of gravity, the inertia tensor and friction parameters. Among the techniques that have been proposed for their determination, the identification of dynamic parameters through experimental methods is the one that has provided better results. However, the dynamic identification of robotic systems is far from being resolved [2].

The parameter identification process consists of fitting measured data to the response of the dynamic model. It is well known that the dynamic model of a robotic system can be written in linear form with respect to the dynamic parameters to be identified, provided that the friction model considered is a linear one. Moreover, depending on the system configuration, not all the dynamic parameters contribute to the robot dynamics; therefore, only a subset of so-called base parameters can be identified. However, when a parameter identification process is performed experimentally, not even base parameters can be correctly identified due to noises in measurements and discrepancies in modeling. Thus, parameters with an independent contribution less relevant than others are difficult to identify. For instance, in the identification of a simulated Puma-like serial robot it has been reported that for 36 base parameters, with a given noise level, only 15 parameters were properly identified [3].

For parallel robots, not all the joints are active and they have a limited workspace compared to serial robots. These facts makes that the excitation of all the base parameters becomes even more difficult when comparing with serial robots. This fact was highlighted in reference [4] where it is set up the need of an iterative methodology that considers what parameters are relevant to be identified. The general idea is that a complete and complex dynamic model does not lead to improve the results when the parameters are not properly identified. Thus, it is preferable to use a reduced model with parameters properly identified.

An approach for finding an appropriate parameterization of the identified model for a 6-DOF robot was presented in [5]. Considerations of large amounts of noise in measurements for the dynamic identification of a 6-DOF robot were presented in [6]. In addition, some practical rules, based on experiments, have been presented for experimental parameter identification of 6-DOF parallel robots [7]. Other authors have implemented bounded error techniques in which the solution is a set of parameter vectors consistent with measurement data, prior error bounds and modeling hypotheses [8]. Recently, a simplified model of a class of parallel robot was developed based on the considered robot structure [9].

Basically, two approaches have been presented for reducing the dynamic model. On the one hand, since the dynamic model can be written in linear form with respect to the parameters to be identified, the well-known linear techniques can be applied to evaluate which parameter are properly identified, for instance the relative standard deviation has been used as a reduction criteria [10]. On the other hand, the contribution of the estimated parameters to the joint torques can be used to guide the reduction process [11]. Regardless of the criteria used for reducing the model, in both approaches, the reduction process stops by means of an empirical criterion. When reduction is carried out through the relative standard deviation, in reference [12], it is proposed to reduce it until the relative standard deviation of the parameter with the maximum value reaches 10 times the one with minimum value. On the other hand, when the reduction is carried out by considering the contribution of the parameter to the joint forces, the threshold is user-defined [11].

In the reduction strategies presented so far, the physical feasibility was not considered which is an aspect of paramount importance given the fact that a physical feasible set of parameters contributes to model-based control or motion simulation of robots [13]. Moreover, this fact is crucial when the values of parameters are needed for physical understanding of the robot characteristics. The identification of parallel robots could lead to identified parameters with unfeasible values. For instance, in the identification of a 6-DOF parallel robot, one of the parameters was unfeasible [14]. This fact was dealt with by eliminating the parameter with the unfeasible value and after that, authors carried out a new identification with the reduced model. The physical feasibility was also considered in the identification of a 3-RPS parallel robot by using non-linear constrained optimization [15], however, a complete and complex model was used, thus, reduction process was not considered. In addition, identification was carried out by nonlinear optimization techniques which can lead to a local minimum. Another important aspect that can be highlighted is that, for fully parallel robots, considerations regarding symmetries and geometries of the robot parts can be used to simplify the dynamic model [16, 17].

In this paper, a methodology to identify the dynamic parameters of parallel robots in terms of relevant parameters is put forward. Its application is experimentally tested on two different configurations of actual 3-DOF parallel robots. The objective of the proposed methodology is to start from a full dynamic model, then simplification concerning the geometry of each link and, the symmetry due to legs of fully parallel robots, are carried out. After that, the identification is done by Weighted Least Squares. Then, with statistical considerations, the model is reduced until the physical feasibility conditions are met. The parameters constituting the reduced model, obtained by the proposed methodology, are called relevant parameters. It is worth noting that, to the author's best knowledge, obtaining the reduced model considering 1) the simplification, due to symmetries and geometries of the robots parts and 2) by using the physical feasibility conditions, has not been proposed before.

The implementation of the proposed methodological strategy allows to find a reduced model that has been verified by a comparison among the generalized forces from the identified models and the measured forces from the experiments. The forward dynamic problem in terms of the identified parameters is also addressed, which is a topic hardly handled in previous papers dealing with dynamic identification. To this end, an approach to write the forward dynamic problem in terms of the identified parameters is put forward.

The paper is organized as follows. In Section 2, the linear dynamic model with respect to the dynamic parameters is developed and the steps of the proposed methodology, including a flowchart for its application, are presented. The results of applying the proposed methodology over a 3-RPS parallel robots are shown in Section 3 and for a 3-PRS robots in Section 4. In Section 5 the forward dynamic in terms of relevant parameters are developed. Finally, the main conclusions are presented.

## 2. Methodology

### 2.1. Dynamic Model

It is well known that the dynamic model of a parallel robot can be written in linear form with respect to the dynamic parameters, and thus, linear estimators can be used. In what follows, the basic equations for developing the dynamic model of parallel robots are presented. For a full understanding of the terms appearing in the equations that follow references are included for the reader.

#### 2.1.1. Rigid Body Model

The equation of motion describing the dynamic behavior of a parallel robot can be obtained by making a cut on one or more of its joints to obtain an equivalent system of several open chains. After that, constraint equations due to the joints cut can be included by means of Lagrange multipliers. The dynamic model of an open chain can be written linearly with respect to the dynamic parameters following the procedures presented in references [12, 18]. Thus the dynamic model, for a joint $k$ and a particular pose $t-th$ of the robot, can be written in matrix form as follows,

$$\vec{\tau}_k = \mathbf{K}_k\left(\vec{q}, \dot{\vec{q}}, \ddot{\vec{q}}\right) \vec{\Phi}_{rb} \qquad (1)$$

where $\tau_k$ is the generalized joint forces. Vector $\vec{\Phi}_{rb}$ regroups the elements of the inertia tensor $[\ I_{xx_i}\ \ I_{xy_i}\ \ I_{xz_i}\ \ I_{yy_i}\ \ I_{yz_i}\ \ I_{zz_i}\ ]^T$, the mass $m_i$ and the first mass moments with respect to the center of gravity $[\ mx_i\ \ my_i\ \ mz_i\ ]^T$. All of them are expressed with respect to a non centroidal local reference system attached to the link. Vectors $\vec{q}, \dot{\vec{q}}, \ddot{\vec{q}}$ are the generalized coordinates and their time derivatives. Matrix $\mathbf{K}_k$ can be built as follows,

$$\mathbf{K}_k = \begin{cases} {}^k\vec{z}_k^T \cdot \sum_{i=k}^{n} {}^k\mathbf{R}_i \left[ {}^i\hat{\eta}_i + \left( {}^i\tilde{r}_{O_k O_i} {}^i\tilde{\eta}_i - {}^i\ddot{\tilde{r}}_{O_i} \right) + {}^i\tilde{r}_{O_k O_i} {}^i\ddot{\vec{r}}_{O_i} \right] & \text{R joint} \\[1em] {}^k\vec{z}_k^T \cdot \sum_{i=k}^{n} {}^k\mathbf{R}_i \cdot \left( {}^i\ddot{\vec{r}}_{O_i} + {}^i\tilde{\eta}_i \right) & \text{P joint} \end{cases} \quad (2)$$

In Eq. 2 all of the kinematic terms are expressed with respect to a local reference frame. The Denavit-Hartenberg modified convention has been considered for modeling the system, ${}^i\hat{\eta}_i = \left( {}^i\dot{\hat{\omega}}_i + {}^i\tilde{\omega}_i {}^i\hat{\omega}_i \right)$, ${}^i\tilde{\eta}_i = \left( {}^i\tilde{\omega}_i {}^i\tilde{\omega}_i + {}^i\dot{\tilde{\omega}}_i \right)$, $\vec{z}_k$ is, as can be seen in Fig. 1, the vector along joint axis $k$, $\vec{r}_{O_k O_i}$ is the position vector of $O_i$ with respect to $O_k$, and $\ddot{\vec{r}}_i$ is the linear acceleration of frame $i$. Script P applied to a prismatic joint and R for a revolution joint, $i = k..n$ indicates the sum over all links above joint $k$, including itself. Matrix $\tilde{a}$ is the skew antisymmetric matrix associated to the cross product vector, and Matrix $\hat{a}$ is an operator defined as follows,

$$\hat{a} = \begin{pmatrix} a_x & a_y & a_z & 0 & 0 & 0 \\ 0 & a_x & 0 & a_y & a_z & 0 \\ 0 & 0 & a_x & 0 & a_y & a_z \end{pmatrix} \quad (3)$$

For a closed-chain system, Lagrange multipliers can be used for including the constraint equations. By doing so, and considering the coordinate partitioning method [19], the dynamic model of a parallel robot, which is linear with respect to the dynamic parameters, can be written as follows,

$$\vec{\tau}_i - \mathbf{X}^T \cdot \vec{\tau}_d = \left[ \mathbf{K}_i - \mathbf{X}^T \cdot \mathbf{K}_d \right] \cdot \vec{\Phi}_{rb} \quad (4)$$

where $\mathbf{X} = \mathbf{A}_d^{-1} \cdot \mathbf{A}_i$. Matrices $\mathbf{A}_d$ and $\mathbf{A}_i$ are obtained when the coordinate partitioning is applied to the Jacobian matrix of the constraint equations. Subscripts $d$ and $i$ stands for dependent an independent generalized coordinates, respectively.

If the parallel robot being studied is non-redundant, Eq. 4 becomes

$$\vec{\tau}_i = \left[ \mathbf{K}_i - \mathbf{X}^T \cdot \mathbf{K}_e \right] \cdot \vec{\Phi}_{rb} = \mathbf{K}_{rb} \cdot \vec{\Phi}_{rb} \quad (5)$$

In addition, in Eq. 5 subscript $rb$ indicates that matrix $\mathbf{K}$ is related to the rigid body base parameters $\vec{\Phi}_{rb}$.

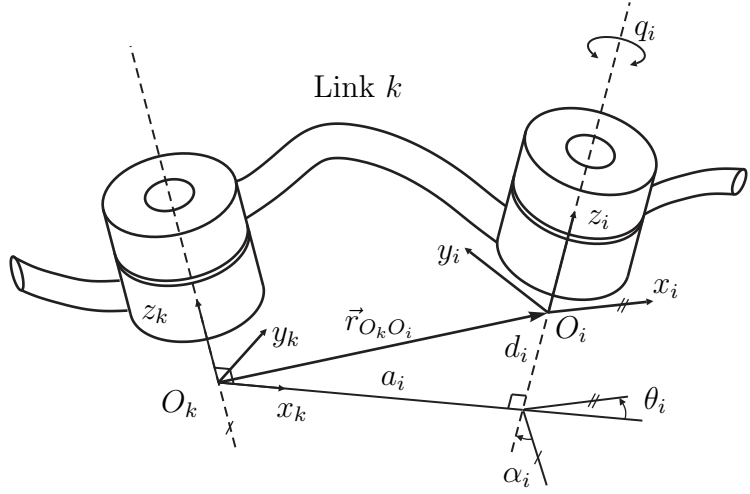

Figure 1: D-H notation for defining link position and orientation

*2.1.2. Friction Models*

Friction is presented in the joints of most of the robotic systems. Several non-linear and linear models has been proposed, see for example [20] among others, but due to its simplicity, a widely implemented model for parameter identification is the linear one in Coulomb and viscous friction parameters. This model is a function of joint velocities [21, 22],

$$F_f = \begin{cases} F_c^+ + F_v^+ \dot{q} & \dot{q} > 0 \\ -F_c^- - F_v^- \dot{q} & \dot{q} < 0 \end{cases} \quad (6)$$

where $F_c$ and $F_v$ stand for the Coulomb and viscous friction coefficients, respectively. Eq. 6 can be applied to passive and active joints. Since the dynamic model was written in joint coordinates, the inclusion of friction is straightforward, as follows

$$\vec{\tau}_{f_i} = \left[\mathbf{K}_{f_i} - \mathbf{X}^T \cdot \mathbf{K}_{f_d}\right] \cdot \vec{\Phi}_f = \mathbf{K}_f \cdot \vec{\Phi}_f \quad (7)$$

where subscript $f$ indicates that matrix $\mathbf{K}_f$ is related to the friction parameters $\vec{\Phi}_f$.

*2.1.3. Active elements*

The actuator dynamics and its driving system is herein referred to as active elements. These elements, as was noted in the experiments conducted, consume a considerable part of the actuator force. Considering the rotor and gear inertia, the corresponding equation for joint $i-th$ can be written as follows,

$$\tau_{r_i} = \left(J_j + \Gamma_i^2 J_{m_i}\right) \ddot{q}_i = J_{r_i} \ddot{q}_i \tag{8}$$

where $J_j$ and $J_{m_i}$ stand for the rotor and driving system inertia, respectively. Parameter $\Gamma_i > 1$ is the gear ratio and subscript $r$ indicates that matrix $\mathbf{K}_r$ is related to the inertia parameters of the driving system $\vec{\Phi}_r$.

If one considers that the active elements are located on the active joints, their inclusion in the dynamic model is done as follows,

$$\vec{\tau}_{r_i} = \mathbf{K}_{r_i} \cdot \vec{\Phi}_r \tag{9}$$

2.1.4. Full robot model

The complete robot model can be built by combining Eqs. 5, 7 and 9, as follows:

$$\vec{\tau}_i = \begin{bmatrix} \mathbf{K}_{rb} & \mathbf{K}_f & \mathbf{K}_r \end{bmatrix} \cdot \begin{bmatrix} \vec{\Phi}_{rb} \\ \vec{\Phi}_f \\ \vec{\Phi}_r \end{bmatrix} = \mathbf{K}\left(\vec{q}, \dot{\vec{q}}, \ddot{\vec{q}}\right) \cdot \vec{\Phi} \tag{10}$$

Eq. 10 is valid when the friction model is a linear one. Nonetheless, when non-linear friction models are included, the complete robot model can be written as follows,

$$\vec{\tau}_i = \begin{bmatrix} \mathbf{K}_{rb} & \mathbf{K}_r \end{bmatrix} \cdot \begin{bmatrix} \vec{\Phi}_{rb} \\ \vec{\Phi}_r \end{bmatrix} + \left(\vec{F}_i - \mathbf{X}^T \cdot \vec{F}_d\right) \tag{11}$$

where $\vec{F}_i$ and $\vec{F}_d$ stand for the independent and dependent non-linear friction models, respectively.

2.2. Base parameters

On the whole, the number of parameters to be identified is greater than the number of equations that, with the procedure followed in the previous section, are equal to the robot DOF. Thus, Eq. 11 is an undetermined system. In order to identify the parameters, an overdetermined system is built by applying Eq. 10 to different poses of the robot sampled while the robot is performing a prescribed trajectory. When the dynamic model is linear with respect to the dynamic parameters, the different equations, that result from the different poses, can be put together to form the following system,

$$\mathbf{W}\left(\vec{q}, \dot{\vec{q}}, \ddot{\vec{q}}\right) \cdot \vec{\Phi} = \vec{Y} \tag{12}$$

where $\mathbf{W}$ is called the observation matrix, and $\vec{Y}$ is the vector collecting the generalized forces of the different robot poses.

As has been mentioned before, the set of parameters that can be identified are the base parameters. For serial robots, symbolic procedures for obtaining the base parameter model have been proposed [12]. However, for parallel robot numerical procedures are often used, for instance, an approach based on singular values decomposition (SVD). When applying the SVD the selection of the set of base parameters is not unique. In the experiments that were conducted on the robots, it was found that the experimental design can be improved by taking into consideration the selected set of base parameters. The topology of the robot, as well as the type of the active joints, affects significantly the values of the condition number of the observation matrix that can be achieved through the design of experiments. Taking into account that the condition number limited the input/output transmission error when using Least Square Methods for identification, in this paper the selection of the set of base parameters is found as follows,

- Determine the possible sets of base parameters. To find the sets of parameters, the SVD procedure proposed by [23] can be implemented.

- For each set of base parameters, run the optimization processes presented in the next section, considering the condition number of the observation matrix as a criterion.

- Select the set of parameters which gives the lowest values according to the criterion used in the step before.

Another important aspect to consider is how the robot is driven. Generally, driving is given by 1) a linear movement through a prismatic joint or 2) an angular movement through a revolution joint. The case in which the robot is driven through linear actuator, a set of base parameters whose physical feasibility is easier to evaluate can be reached. As an example, consider the fact that the active joints could be prismatics. Since the mass of the link connected through the joint contributes to the forces being applied to the prismatic joint, a possible formulation of the base parameter is a parameter including the mass of the link grouped with the link masses over the actuated joint. In detail, consider the 3-PRS depicted in Fig. 2. Robot contains 7 links, hence rigid body model contains 70 parameters. Nonetheless, only 24 parameters contribute to the dynamics which are grouped into 19 base parameters. A possible base parameters set for the 3-PRS robot is listed in Table 1.

Table 1: Base parameters for the 3-PRS parallel robot.

| No | Base Parameter | No | Base Parameter |
|---|---|---|---|
| 1 | $I_{zz_2} - lr^2 \sum_{i=1}^{2} m_i$ | 11* | $my_3 - ly_{p_3} \sum_{i=1}^{5} m_i$ |
| 2 | $mx_2 - lr \sum_{i=1}^{2} m_i$ | 12 | $mz_3$ |
| 3 | $my_2$ | 13 | $I_{zz_5} - lr^2 \, m_{4,5}$ |
| 4 | $I_{xx_3} - ly_{p_3}^2 \sum_{i=1}^{5} m_i$ | 14 | $mx_5 - lr \sum_{i=4}^{5} m_i$ |
| 5 | $I_{xy_3} + lx_{p_3}ly_{p_3} \sum_{i=1}^{5} m_i$ | 15 | $my_5$ |
| 6 | $I_{xz_3}$ | 16 | $I_{zz_7} + lr^2 \sum_{i=1}^{5} m_i$ |
| 7 | $I_{yy_3} - lx_{p_3}^2 \sum_{i=1}^{5} m_i + lx_{p_2}^2 \, m_{4,5}$ | 17* | $\sum_{i=1}^{7} m_i$ |
| 8 | $I_{yz_3}$ | | |
| 9 | $I_{zz_3} - lx_{p_2}^2 \sum_{i=1}^{3} m_i$ | 18 | $mx_7 + lr \sum_{i=1}^{5} m_i$ |
| 10 | $mx_3 - lx_{p_3} \sum_{i=1}^{5} m_i + lx_{p_2} \sum_{i=4}^{5} m_i$ | 19 | $my_7$ |

As can be seen from the table, masses $m_1$ to $m_6$ were considered as parameters for linear combination. Thus, parameter $\Phi_{17}$ results as the sum of all link masses. In this way, one can verify the physical feasibility of the $\Phi_{17}$ whose values must be positive.

*2.3. Identification method*

The parameter estimation of $\vec{\Phi}$ can be achieved by solving the overdetermined linear systems through Weighted Least Square (WLS). This method is the most widely implemented technique and leads to an improvement of results. The weighted dynamic model can be express as,

$$\mathbf{W}_w \vec{\Phi}_{base} = \vec{\tau}_w + \vec{\rho}_w \tag{13}$$

where $\mathbf{W}_w = \Sigma \, \mathbf{W}$, $\tau_w = \Sigma \, \vec{\tau}$ and $\rho_w = \Sigma \, \vec{\rho}$. The vector $\vec{\rho}$ stands for unmodelled dynamic of actuators and measurement noise.

The WLS solution $\vec{\Phi}_{base}$ minimizes the two norms of the vector of weighted errors $\rho_w$. If the identified model is linear, with respect to the parameters, vector $\vec{\Phi}_{base}$ can be obtained as follows,

$$\vec{\Phi}_{base} = \left(\mathbf{W}_w^T \mathbf{W}_w\right)^{-1} \mathbf{W}_w^T \vec{\tau} \qquad (14)$$

The weighted matrix can be built as a diagonal matrix containing the estimated variance of each measured actuator forces, which can be estimated as,

$$\sigma_i^2 = \frac{1}{n_{pts}(M-1)} \sum_{k=1}^{n_{pts}} \sum_{j=1}^{M} (\tau_{ij}(k) - \overline{\tau}_i(k))^2 \qquad (15)$$

where $\tau_{ij}(k)$ represents the $k-th$ measurement of the torque at joint $i$ during the $j-th$ trajectory repetition and $\overline{\tau}_i$. represents the average over the $M$ repetitions of the value of the $i-th$ joint torque at the $k-th$ measurement. The diagonal matrix $\Sigma$ is built with values of the inverse of the estimated variance as $\Sigma = \mathrm{diag}[1/\sigma_1, \ldots, 1/\sigma_n]$.

2.4. Design of trajectories

As mentioned before, Eq. 12 is obtained by sampling data along a given trajectory performed by the robot. Thus, an important step in the identification process is the selection of the trajectory to be performed by the robot. The design of trajectories can be seen as an optimization process where a parameterized trajectory is optimized considering as an objective function a criterion for improving the excitation of the dynamic parameters [24]. Constraint equations due to the joints movements and reachable robot workspace are included as restrictions in the optimization process. Taking into account the fact that the variance of the forces is estimated through repetitions of the trajectory, the following parameterization presented in [25] is suitable for identification purposes,

$$q_i(t) = q_{i0} + \sum_{j=1}^{n_H} \left[\frac{a_{ij}}{2\pi f \cdot j} \sin(2\pi f \cdot j \cdot t) - \frac{b_{ij}}{2\pi f \cdot j} \cos(2\pi f \cdot j \cdot t)\right] \qquad (16)$$

where $t$ is the time, $q_{i0}$, $a_{ij}$ and $b_{ij}$ are the coefficients of the Fourier series that will be the variables of the optimization process, $n_H$ is the harmonic number and $f$ is the fundamental frequency.

Here it is proposed to parameterize the trajectory in terms of joint coordinates. For 3-DOF parallel robots the solution of the direct kinematic problem is not as cumbersome as for a 6-DOF. In addition, in joint coordinates constraint equations due to the actuator movements can be easily introduced in the optimization problem since in these coordinates they appear as linear.

Let us define $\vec{\delta}_p$ as the vector that grouped the coefficients of the above-defined Fourier series. The optimization problem can then be seen as a non-linear optimization problem subjected to non-linear constraints. This can be written as follows,

$$\begin{aligned} \text{minimize } f(\vec{\delta}_p) & \quad \delta_p \in R^{n_p} \\ \text{subject to : } & \quad c(\vec{\delta}_p) \leq 0 \\ & \quad \mathbf{A} \, \vec{\delta}_p \leq b \end{aligned} \quad (17)$$

where $f(\vec{\delta}_p)$ is an objective function. Two well-accepted criteria for choosing the objectives function are: 1) The condition number of the observation matrix $\kappa(\mathbf{W})$ [24], 2) To maximize a scalar measurement of the information matrix, $-log\left(det\left(W^T \Sigma^{-1} W\right)\right)$ [26]. For the proposed methodology the condition number of the observation matrix $\kappa(\mathbf{W})$ was selected as the objective function. The optimization process can be summarized as follows,

- Parameterization of the joint independent coordinates by means of the Fourier series proposed in [25].

- Defining system constraints due to the joint movements and reachable workspace.

- Selection of the criterion for optimization.

- Optimization by non-linear techniques such as Sequential Quadratic Programing (SQP).

*2.5. Model reduction*

The objective of the proposed methodology is to find a model capable of properly estimate the system dynamic response. Moreover, the inherent parameters of the identified model must be physically feasible. The base parameters are the identifiable parameters, but not even base parameters can be correctly identified since some of them have insignificant contribution to the system dynamic response or the trajectory being used for the identification do not excites properly these parameters. These parameters are prone to be corrupted by noises and unmodelled dynamics. In order to improve the noise immunity and reduce the computational burden of the dynamic models, a simplified model that properly describes the systems response can be found. Two simplification can be done:

- Simplification with regard of the geometry of the links and due to the symmetries presented in parallel robots.

- The elimination of parameters that are not properly identified.

As an example, consider the case of the 3-PRS parallel depicted in Fig. 2. Due to the geometry of the coupler, links 2, 5 and 7, it can be assumed that the center of gravity lies in the axis normal to the cross section of the link, thus some parameters related to the location of the gravity center have their value equal or near to zero. The form of the platform, circular and flat, can also be used for simplification.

The symmetries of links allow us to make simplifications in the number of parameters. For instance, the masses of links 1, 4 and 6 can be assumed to be equal. The same assumption can be done over link 2, 5 and 7. It is worth noting that, even though the geometry of legs are slightly different, due to the inevitable manufacturing tolerances, the effect of reducing the number of parameter of the model improves the noise immunity in such a way that the manufacturing tolerances are not as relevant as the reduction of the model. Further, it is shown that the reduced model is closer to the experiments, when comparing to the complete model, not only when solving the inverse dynamic problem, but also when solving the forward dynamics.

To summarize, previously to the identification process the analyst must consider the reduction of the model by observing the geometry of the robot, as well as the symmetries presented by the characteristics of the parallel robot.

In Table 2 are listed the base parameters model considering the simplification due to the link geometries and symmetries. As can be seen, the model contains 13 rigid body parameters.

On the other hand, the next step to find the reduced model is to eliminate those parameters that are not properly identified.

Table 2: Simplified Model for the 3-PRS parallel robot.

| No | Base Parameter |
| --- | --- |
| 1 | $I_{xx_3} - ly_{p_3}^2 \ (2m_1 + 2m_2 + m_3)$ |
| 2 | $I_{xy_3} + lx_{p_3}ly_{p_3} \ (2m_1 + 2m_2 + m_3)$ |
| 3 | $I_{xz_3}$ |
| 4 | $I_{yy_3} - lx_{p_3}^2 \ (2m_1 + 2m_2 + m_3) + lx_{p_2}^2 \ (m_1 + m_2)$ |
| 5 | $I_{zz_3} - lx_{p_2}^2 \ (m_1 + m_2 + m_3)$ |
| 6 | $mx_3 - lx_{p_3} \ (2m_1 + 2m_2 + m_3) + lx_{p_2} \ (m_1 + m_2)$ |
| 7 | $my_3 - ly_{p_3} \ (2m_1 + 2m_2 + m_3)$ |
| 8 | $mz_3$ |
| 9 | $I_{zz_2} - lr^2 \ (m_1 + m_2)$ |
| 10 | $mx_2 - lr \ (m_1 + m_2)$ |
| 11 | $I_{zz_2} + lr^2 \ (2m_1 + 2m_2 + m_3)$ |
| 12 | $(3m_1 + 3m_2 + m_3)$ |
| 13 | $mx_7 + lr \ (2m_1 + 2m_2 + m_3)$ |

*2.5.1. Stepwise identification*

After having reduced the model by considering link geometries and the symmetries of the robot, the model is reduced through the evaluation of the relative standard deviation of the estimated parameters. The reduction can be done as mentioned in [12]. The parameter with the maximum standard deviation is eliminated from the model. A distinctive feature of our criterion, with respect to the criterion presented in [12] and [11], is the way we stop the reduction process and hence one of the contributions this paper. In the proposed methodology the stopping criterion for model reduction is established by considering the physical feasibility conditions.

The identification starts with the use of a complete and complex dynamic model. Then, the estimation of the dynamics parameters is carried out by means of the WLS, linear techniques are used provided that the model is linear. After identification the feasibility conditions are verified. If the set of parameters are not feasible, then, reduction continues until the set of parameters corresponding to a reduced model is physically feasible. This is summarized in Algorithm 1. In the algorithm $r$ are $c$ are the rows and columns of the observation matrix $\mathbf{W}$.

It can been seen in Algorithm 1 that the reduction process is halted

```
while flag=false do
    Find $\mathbf{W}_w = \Sigma\ \mathbf{W}$ and $\vec{\tau}_w = \Sigma\ \vec{\tau}$;
    Find $\vec{\Phi} = \left(\mathbf{W}_w^T\mathbf{W}_w\right)^{-1}\mathbf{W}_w^T\vec{\tau}_w$;
    Find relative standard deviation;
    $s^2 = \dfrac{\|\vec{\tau}_w - \mathbf{W}_w\Phi\|}{(r-c)}$;
    $\mathbf{C}_\Phi = s^2\ (\mathbf{W}_w^T\mathbf{W}_w)^{-1}$;
    for $i \leftarrow$ to $length(\mathbf{C}_\Phi(:,1))$ do
        $\sigma_i = \dfrac{\mathbf{C}_{\Phi_{ii}}}{\Phi_i}100$
    end
    $j = max\left(\vec{\sigma}\right)$;
    Reduction $\Phi_i$;
    run feasibly algorithm;
    if flag=true then
        feasible=true
    end
end
```

Algorithm 1. Reduction process

when a feasible set of parameters is obtained. The way of verify if a set of parameters is physically feasible is presented in the following section.

*2.6. Physical feasibility*

Physical feasibility refers to the condition in which a set of identified parameters assure the fact that the kinetic energy of the system must be positive,

$$m_i > 0 \tag{18}$$
$$^i\mathbf{I}_{G_i} > 0$$

As can be seen in Eq. 18, the condition is established in terms of masses and inertia tensors expressed with respect to the center of gravity. However, the identified parameters are the base parameters, which are linear combination of the link parameters. Thus, the feasibility equations have to be rewritten in terms of the link parameters. First, the relation between the base parameters and the link parameters can be written by expressing the base parameter combination in matrix form,

$$\vec{\Phi}_{base} = \vec{\Phi}_1 + \mathbf{B}\, \vec{\Phi}_2 \tag{19}$$

where $\vec{\Phi}_{base}$ are the identified parameters, $\vec{\Phi}_1$ and $\vec{\Phi}_2$ are the link parameters. Matrix $\mathbf{B}$ contains the linear relations between link parameters and base parameters.

From Eq. 19 vector $\vec{\Phi}_1$ can be written as a function of the identified base parameters and some of the link parameters grouped in vector $\vec{\Phi}_2$,

$$\vec{\Phi}_1 = \vec{\Phi}_{base} - \mathbf{B}\, \vec{\Phi}_2 = f\left(\vec{\Phi}_{base}, \vec{\Phi}_2\right) \tag{20}$$

After obtaining the base parameters through identification and by giving values to the parameters grouped in vector $\vec{\Phi}_2$, one can find the remain link parameters and hence verify the feasibility. Physical feasibility is verified by writing Eq. 18 in terms of link gravity center parameters by means of the Steiner theorem.

Feasibility is used in this paper as a flag for determining the set of relevant parameters. The stepwise identification is applied to the reduced model until the flag of physical feasibility is achieved. In order to find the link parameters $\vec{\Phi}_1$, a search space for $\vec{\Phi}_2$ is defined, for instance $lb \leq \vec{\Phi}_2 \leq ub$. This search space is then discretized and $\vec{\Phi}_1$ is found through all the discrete space. If a set of feasible parameter satisfying Eq. 18 can be found, the reduction process stops. As an example of the verification of the feasibility condition, consider the set of parameters in Table 2 where the values of link parameters can be found, starting from the identified values of the base parameters, by assigning values to the masses $m_1$, $m_2$ and $m_3$. Moreover, $m_3$ can be obtained from base parameter 12 and masses $m_1$ and $m_2$. Thus, link parameters can be found with the base parameter solution and assigned values of $m_1$ and $m_2$. The verification of feasibility for this particular case is summarized in Algorithm 2.

The strategy presented in this section, based on the steps presented so far, is summarized by means of a flowchart in Fig. 3. As can be seen in the flowchart, the identification is carried out by means of linear techniques as WLS. Having applying WLS the obtained values are handled through a process, algorithm 2, in order to verify whether the set of identified parameters is physically feasible or not. If the set of identified parameters is unfeasible, the reduction process continues.

```
Input: Φ_base, ub_1, lb_1, ub_2, lb_2, np
for i = lb_1 : (ub_1 − lb_1)/np : ub_1 do
    for i = lb_1 : (ub_2 − lb_2)/np : ub_2 do
        m_1 = i ;
        m_2 = j ;
        Φ⃗_1 = Φ⃗_base − BΦ⃗_2 ;
        ^iI_{G_i} = f (Φ_1, Φ_2) ;
        m_i = f (Φ_1, Φ_2) ;
        if ^iI_{G_i} > 0 y m_i > 0 then
            flag= 1
        end
    end
end
```

Algorithm 2. Checking physical feasibility

It is important to mention that, in some cases the identification might not reach a set of feasible parameters. These cases could happen when the starting complete and complex model does not describe in an appropriate manner the dynamic response of the system being identified. For instance, we have started the identification of a 3-RPS robot omitting the inertial of the active elements. As a result, we have found a set of parameter which were properly identified in terms of the statistical criteria, but with parameters containing non feasible values [17].

## 3. Application to the 3-PRS robot

The proposed methodology is implemented for the dynamic parameter identification of two fully parallel robots with 3-RPS and 3-PRS configurations. Fig. 4 shows the robots considered. First the application for the identification of the 3-PRS is presented. This robot has been used as an example throughout the development of the methodology.

In order to implement the trajectories performed by the robot, a control systems were developed. An industrial PC has been used. The industrial PC is equipped with 2 data acquisition cards. A card Advantech PCI-1720 has been used for supplying the control actions; and a PCI-833 has been used for reading encoders. The proposed control architecture gives two advantages, first the simplicity and second the low cost. In addition, this open architec-ture gives a powerful platform for programming high level tasks. Because the control unit is based on PC, it is a totally open system. So any controller and/or control technique can be programmed and implemented such automatic trajectory generation, control based on external sensing using force sensor or artificial vision, cooperative control of several robots. In this work, the Microsoft Visual C++ environment has been employed to implement a PID control. In addition, the PC environment allows us to connect the data structure with commercial CADCS packages.

*3.1. Experimental Setup*

- Optimal trajectories where executed by the robot by means of a PID control.

- Each trajectory was repeated 5 times during the experiments and the controls action were applied at 100Hz. The total duration of each trajectory was 8 s.

- The motor positions were measured from encoders at a sample rate of 100 Hz. The repetition of the trajectory allows us the estimation of the position deviation, which for each actuator was: 1.0628e-002, 7.2504e-003 y 9.1636e-003 mm. This fact indicated that the assumption of free noise in input variables can be applied.

- The forces used for identification were found by establishing a linear relation between controls action and forces. This relation was verified through experiments that were conducted previously on motors.

- Identification is carried out using WLS.

Since the measurement of the joint velocities and accelerations require special devices, their determination is carried out by indirect approach. Generally, two approaches have been used for their determination:. 1) Numerical differentiation techniques which require the design of effective filters [12] or 2) Calculation of velocities and accelerations by way of the measured positions, in an analytical way [26]. In the later approach, the measured position is fitted to the Fourier Series, Eq. 16, with the number of harmonic function similar to the one used for trajectory optimization.

## 3.2. Identification by the proposed methodology

The complete model included the 13 rigid body parameters listed in Table 2, 12 friction parameters which considered linear Coulomb and viscous friction for the prismatic and revolution joints, and 3 parameters regarding the actuator dynamics. Thus, the reduction process started with 28 parameters. Then the proposed methodology was applied. The process finalized with a model containing 13 parameters corresponding to: 4 rigid body parameters, 6 friction parameters for modeling friction in prismatic joints, and 3 parameters for actuator dynamics. The values of the parameters of the reduced model, along with the relative standard deviation, are listed in Table 3.

Table 3: Identified Model for the 3-PRS [SI Unit]

|  | $\Phi_i$ | $\sigma_i$ |
|---|---:|---:|
| $I_{xx_3} - ly_{p_3}^2 \left(2m_1 + 2m_2 + m_3\right)$ | -11.57 | 8.22 |
| $I_{xy_3} + lx_{p_3} ly_{p_3} \left(2m_1 + 2m_2 + m_3\right)$ | 7.82 | 6.48 |
| $my_3 - ly_{p_3} \left(2m_1 + 2m_2 + m_3\right)$ | -18.76 | 0.726 |
| $3m_1 + 3m_2 + m_3$ | 62.78 | 0.596 |
| $Fv_1$ | 3509.89 | 3.92 |
| $Fc_1$ | 100.16 | 3.97 |
| $Fv_2$ | 3418.03 | 3.75 |
| $Fc_2$ | 122.46 | 1.49 |
| $Fv_3$ | 3870.60 | 4.69 |
| $Fc_3$ | 100.64 | 1.39 |
| $Jr_1$ | $1.7078 \cdot 10^{-3}$ | 3.38 |
| $Jr_2$ | $1.5334 \cdot 10^{-3}$ | 1.16 |
| $Jr_3$ | $1.6201 \cdot 10^{-3}$ | 4.51 |

## 3.3. Validation

Table 5 shows the results for validation of the complete and reduced model. The response of the inverse dynamic model was evaluated by the mean relative error $\varepsilon_{ra}$ for 5 trajectories which were different from the ones used in identification. It is clearly seen that, the reduced model has improved the results compared to the full model not only in terms of the mean values for different trajectories, but also when evaluating the deviation from the mean.

Table 4: Mean and deviation of $\varepsilon_{ra}$ for validation trayectories 3-PRS tobot.

|  | Mean | Deviation |
|---|---|---|
| Complete Model | 16,7 | 2,997 |
| Reduced Model | 14,9 | 1,22 |

A qualitative validation of the inverse dynamic solution for the reduced model has been performed and the results are very similar. As can be seen in Fig. 5, the estimated forces follows the measured forces very closely.

## 4. Application to the 3-RPS robot

The identifiability of the dynamic parameter for the 3-RPS robot used in the experimental framework of this paper was previously performed in [17]. However, in that paper a simulated robot was necessary to carry out the reduction process. Here, it is expected that the application of the proposed methodology will lead to similar results, but in this case the need for the simulated robot is avoided.

The framework for the experiment design and the measured variables is similar to that used for the identification of the 3-PRS robots. The complete model in this case includes 9 rigid body parameters listed in Table 5, and 12 friction parameters which considered linear Coulomb and viscous friction for the prismatic and revolution joints, and 3 parameters regarding the actuator dynamics. Thus, the reduction process started with 24 parameters.

Table 5: Base parameter for the complete robot model of the 3-RPS robot

| No | Base Parameters |
|---|---|
| 1 | $my_1$ |
| 2 | $I_{zz_1} + I_{yy_2}$ |
| 3 | $mz_2$ |
| 4 | $I_{xx_3} - ly_{p_3}\, my_3$ |
| 5 | $I_{yy_3} + ly_{p_3}\, my_3 - lx_{p_2}^2 (m_3 + m_2)$ |
| 6 | $I_{zz_3} - ly_{p_3}^2 (m_3 + m_2)$ |
| 7* | $mx_3 + lx_{p_3}/ly_{p_3} my_3 - lx_{p_2}(m_3 + m_2)$ |
| 8* | $2m_2 - my_3/ly_{p_3} + m_3$ |
| 9* | $m_2 + my_3/ly_{p_3}$ |

The proposed methodology was applied. The process ended with a model containing 12 parameters corresponding to: 3 rigid body parameters, 6 friction parameters for modeling friction in prismatic joints, and 3 parameters for actuator dynamics. The values of the parameters of the reduced model, along with the relative standard deviation, are listed in Table 6.

Table 6: Identified model 3-RPS [SI Unit]

|  | $\Phi_i$ | $\sigma_i$ |
|---|---:|---:|
| $mx_3 + lx_{p_3}/ly_{p_3} my_3 - lx_{p_2}(m_3 + m_2)$ | -2.31 | 6.48 |
| $2m_2 - my_3/ly_{p_3} + m_3$ | 11.18 | 4.09 |
| $m_2 + my_3/ly_{p_3}$ | 6.10 | 2.52 |
| $Fv_1$ | 3489.28 | 3.41 |
| $Fc_1$ | 152.33 | 6.33 |
| $Fv_2$ | 2099.86 | 5.08 |
| $Fc_2$ | 119.94 | 7.00 |
| $Fv_3$ | 3057.99 | 1.87 |
| $Fc_3$ | 202.21 | 1.85 |
| $Jr_1$ | $3.1257 \cdot 10^{-4}$ | 2.04 |
| $Jr_2$ | $2.8695 \cdot 10^{-4}$ | 3.03 |
| $Jr_3$ | $3.4239 \cdot 10^{-4}$ | 0.76 |

The fact that both robots have been built using actuators from the same manufacturer allows us comparisons to be made with respect to the parameters of friction and inertia of the rotors for both robots. The actuators of the 3-PRS robot are more robust than the 3-RPS. The dimensions of the former are higher than the second. This suggests that the inertia of the rotors of the 3-PRS robot must be greater than the 3-RPS. This can be seen in Tables 3 and 6, which verify that the parameters of the robot 3-PRS are greater than the 3-RPS for the three actuators. With respect to the parameters of viscous friction it is not expected that they will show higher discrepancies. Results show that viscous friction parameters are similar in order of magnitude, but the 3-PRS have slightly higher values than those of 3-RPS. Moreover, when comparing the parameters of Coulomb friction they are of the same order of magnitude. Although the PRS robot is expected to present higher values, mainly because the maximum workload of the actuators is greater in the 3-RPS it should be mentioned that the PRS actuators were previously lubri-cated prior to experiments, thus, a direct relationship cannot be established between both robots.

From the study of the reduced models obtained for the parallel robots used in the experiments, it can been seen that parameters corresponding to inertia tensors and moments of first order of robot legs have very little influence on the dynamics for both parallel robots, compared to the parameters of the mobile platform. This is because the type of robot configuration limited the design trajectories that allow the dynamic parameters associated to the legs to become sufficiently excited, so that their inertia properties become relevant on the dynamics. The base parameter that contains the masses of the links and the platform inertia are the identifiable parameters.

On the other hand, both robots present high friction in the prismatic joints, which makes the identification of the rigid body parameters difficult. However, due to the robust design of the 3-PRS robot compared to the RPS robots, more inertial parameters were identifiable. Moreover, it was noted that the configuration of the robot affects the numerical conditioning of the observation matrix. In this regard, the trajectories obtained under the condition number criterion for the 3-RPS robot were of a higher value (between 500-700 order) than that obtained for a 3-PRS configuration (order between 80-200), this result being independent of the magnitude of the dynamic parameters. The latter robot has a better numerical conditioning which can allow us to identify the parameters with greater certainty that the robot RPS. Indeed, we suggest here that this result could be considered when selecting the topology of a robot for an specific application. Since both robots presented allow to control the height and direction of two axes with respect to a local reference system attached to the end-effector of the mobile platform, for the selection of the robot for a particular application one can take into account that the PRS presents better conditioning for identification than the RPS robots. In this hypothetical case, the selection of the PRS instead of the RPS configuration would be preferred. It suggests that the configuration allow us the identification of its parameters with a greater level of certainty, so that control could be improved by means of robust control algorithms. In this respect, it is necessary to continue studies in which both robots are designed to accomplish similar tasks, and then compare the identifiability of the two robots. Finally, the validation results from identified models of the 3-RPS robots are presented.

*4.1. Validation*

Fig. 6 shows the comparison between the predicted forces using the reduced model and the measured forces. It can be seen that both curves are similar.

In order to see how the values of the estimated parameters in the reduction process have changed, Table 7 is included. In the table, the values from a CAD model, which were obtained in such a way that its dynamic response was closed to the actual one [17], is presented. In addition, the values for a complete model, which do not include any simplification or reduction, is included. Moreover, the values for the reduced model have also been included.

Table 7: Some of the identified parameters for the 3-RPS robot [SI Unit]

| $\Phi_i$ | CAD | Reduced Model | Complete Model |
|---|---|---|---|
| $mx_3 + lx_{p_3}/ly_{p_3} my_3 - lx_{p_2}(m_3 + m_2)$ | -2.47 | -2.31 | 1.16 |
| $2m_2 - my_3/ly_{p_3} + m_3$ | 10.83 | 11.18 | -3.29 |
| $m_2 + my_3/ly_{p_3}$ | 5.42 | 6.10 | -0.557 |

As can be observed, the identified parameters of the reduced model and the CAD values, are comparables. The contrary occurred when comparing with those identified using the complete and complex model where significant differences appears. Moreover the set of parameters of the complex model lead to an unfeasible set of parameters.

## 5. Forward dynamic in terms of relevant parameters

The response in terms of the forward dynamic problem was evaluated. To this end, it was necessary to built the forward dynamic in terms of the relevant parameters,

$$\vec{\tau}_i = \mathbf{M}\,\ddot{\vec{q}}_i + \vec{C} + \vec{G} + \left(\vec{F}_{f_i} + \mathbf{X}^T \vec{F}_{f_d}\right) + \mathbf{J}_m\,\ddot{\vec{q}}_i \qquad (21)$$

where $\vec{F}_{f_i}$ y $\vec{F}_{f_d}$ are the vector corresponding to the friction relevant parameters. Subscript $i$ and $d$ apply to the independent and dependent generalized coordinates. Matrix $\mathbf{J}_m$ is a diagonal matrix including the inertial effect of the actuators.

In order to obtain the mass matrix $\mathbf{M}$, the vector corresponding to the centrifugal and Coriolis forces $\vec{C}$ and gravity vector $\vec{G}$, in a suitable for the forward dynamic problem, let us start by written the rigid body dynamic model linear with respect to the dynamic parameters,

$$\vec{\tau_i} = \left(\mathbf{K}_i + \mathbf{X}^T\,\mathbf{K}_d\right)\,\vec{\Phi} \tag{22}$$

In Eq. 22, $\mathbf{K}_i$ is a matrix of size $n \times np$, where $n$ is the number of DOF and $np$ is the number of relevant parameters, $\mathbf{K}_d$ is a matrix of size $(m-n) \times np$, where $m$ is the number of generalized coordinates. Matrix $\mathbf{X}^T$ is of $n \times (m-n)$.

5.1. Mass Matrix

The mass matrix is built based on its properties. That is to say, this matrix contains some of the relevant parameters and the generalized coordinates. Thus, if one set to zero the vector of velocities and gravity, the resulting equation contains the components of the mass matrix. Let us start by expressing the mass matrix in independent and dependent coordinates,

$$\mathbf{M}_{i_{n\times n}}\ddot{\vec{q}}_i + \mathbf{M}_{d_{n\times m-n}}\ddot{\vec{q}}_d \tag{23}$$

The part of the matrix in independent coordinates can be obtained starting from Eq. 22 as follows,

**Input**: $\dot{\vec{q}} = 0$, $\vec{g} = 0$, $\ddot{\vec{q}}_d = 0$
**for** $i=1{:}n$ **do**
    $\ddot{\vec{q}}_i(1:n) = 0$ ;
    $\ddot{\vec{q}}_i(i) = 1$ ;
    $\mathbf{M}_i(1:3,i) = \left(\mathbf{K}_i + \mathbf{X}^T\,\mathbf{K}_d\right)\,\vec{\Phi}$ ;
**end**

The mass matrix corresponding to the generalized coordinates is found as follows,

In order to built the dynamic problem in a state-state configuration the component of dependent accelerations has to be written with respect to the independent accelerations,

$$\ddot{\vec{q}}_d = \mathbf{A}_d^{-1}\vec{b} - \mathbf{X}\cdot\ddot{\vec{q}}_i \tag{24}$$

**Input**: $\dot{\vec{q}} = 0$, $\vec{g} = 0$, $\ddot{\vec{q}}_i = 0$;
**for** *i=1:m-n* **do**
    $\ddot{\vec{q}}_d(1:m-n) = 0$ ;
    $\ddot{\vec{q}}_d(i) = 1$ ;
    $\mathbf{M}_d(1:3,i) = \left(\mathbf{K}_i + \mathbf{X}^T\,\mathbf{K}_d\right)\,\vec{\Phi}$ ;
**end**

By substituting Eq. 24 into Eq. 23 the following equation is obtained,

$$\mathbf{M}_{i_{n \times n}} \ddot{\vec{q}}_i + \mathbf{M}_{d_{n \times m-n}} \ddot{\vec{q}}_d = \mathbf{M}_i \ddot{\vec{q}}_i + \mathbf{M}_d \left[\mathbf{A}_d^{-1}\vec{b} - \mathbf{X} \cdot \ddot{\vec{q}}_i\right] \qquad (25)$$

As can be seen, when replacing the dependent acceleration the term $\mathbf{M}_d \mathbf{A}_d^{-1} \vec{b}$ appears. This matrix depend on the vector of velocities and generalized coordinates, thus, this terms is included in the vector of centrifugal and Coriolis forces $\vec{C}$. The mass matrix is built as follows,

$$\mathbf{M} = \mathbf{M}_i - \mathbf{M}_d \mathbf{X} \qquad (26)$$

*5.2. Vector of Velocity components*

The vector corresponding to the centrifugal and Coriolis forces ($\vec{C}$) is found by the fact that this vector contains the component associated to the vector of generalized coordinates and velocities, this can be written as follows,

**Input**: $\vec{g} = 0$, $\ddot{\vec{q}} = 0$
$C_d(1:n,1) = \mathbf{M}_d \mathbf{A}_d^{-1} \vec{b}$;
$C_i(1:n,1) = \left(\mathbf{K}_i + \mathbf{X}^T\,\mathbf{K}_d\right)\,\vec{\Phi}$ ;
$\vec{C} = C_i + C_d$ ;

where $C_d$ is defined as $\mathbf{M}_d \mathbf{A}_d^{-1} \vec{b}$,

*5.3. Vector of Gravity components*

The vectors of gravity components contains only the generalized coordinates and the vector of gravity, thus, this vector can be found as follows,

**Input**: $\dot{\vec{q}} = 0$, $\ddot{\vec{q}} = 0$
$G(1:n,1) = \left(\mathbf{K}_i + \mathbf{X}^T \mathbf{K}_d\right) \vec{\Phi}$ ;

*5.4. Forward dynamic for the 3-PRS robot*

A simulation was carried out in the Matlab environment using the *ode45* function, which is based on an explicit Runge-Kutta 45 formula. It is a one-step solver that only needs the solution from the immediately preceding time point. It is important to mention that the identified friction model introduces a discontinuity in the equation of motions. Thus, a procedure for solving the discontinuity problem was developed [28]. This procedure consists of the application of an external switch that divides the integration interval into subintervals, the calculation of the friction force in the stick phase is considered equal to the net forces acting in the opposite direction. These procedures can be applied when the integration routine has, as the *ode45* does, the ability to detect an event. Thus, *ode45* was suitable for the purpose of solving the forward dynamic problem.

Table 8 shows the results for validation of the complete and reduced model when solving the forward dynamic problem. The results correspond to the mean values of the relative absolute error between the measured position and the position obtained through simulation, for 10 trajectories which were different from the ones used in identification. Both the mean and deviation values of the prediction error of the generalized coordinates at position level for the reduced model are lower to the ones obtained for the complete model.

Table 8: Mean and deviation of $\varepsilon_{ra}$ in the simulation of the 3-PRS robot.

|  | Mean | Deviation |
|---|---|---|
| Complete Model | 25,2 | 8,679 |
| Reduced Model | 21,5 | 3,78 |

Fig. 7 shows the evolution of the generalized independent coordinates for the complete model, the reduced model and the measured actuator position. As can been seen, the reduced model follows the measured actuator positions more closely than the complete model. Moreover, for some of the validation trajectories, the complete model was not able to accomplish the simulation due to the fact that the evolution in the simulation goes through a singular configuration. This fact could be explained since the complete identified model is prone to being affected by noise and unmodelled dynamic compared to the reduced model.

*5.5. Forward dynamic for the 3-RPS robot*

Fig. 8 shows the comparison between the measured position and the simulations obtained by using the complete model and the reduced model, where it can be seen that the reduced model is closer to the measured position than the complete model.

The verification for 10 different trajectories was also performed. In both cases comparing the measured values with the response of the reduced model and the complete model,the forward dynamic problem, the reduced model was found to predict the response of the system with more accuracy.

The results of validation of the reduced model and the complete and complex model have shown that the former predicts in a more appropriate way the system behavior not only when using the identified model in the solution of the inverse dynamic problem, but also when solving the forward dynamic problem. In addition, for some of the several trajectories used for validation, the complete model failed to achieve the simulation time, indicating that the reduced model whose parameters are physically feasible has advantages over the use of complex models. Moreover, in those cases when the model complete model and the reduced model have accomplished the simulation, the latter was about 20% faster.

On the other hand, one of the advantages of obtaining the forward dynamic with respect to the relevant parameters is that it can be used for model-based control that takes into account the quantification of different terms appearing in Eq. 21.

## 6. Conclusion

On the basis of the experiments that were conducted, it has been found that the use of a complete and complex dynamic model does not always lead to a realistic identification of its dynamic parameters. That is, it has been found that it is preferable to consider simplified models, but with parameters determined with more significance, to define the dynamic behavior of the mechanical system. To this end, in this paper a methodology for the identification of dynamic parameters for parallel robots based on a set of relevant parameters has been proposed. The proposed methodology starts from a complete and complex dynamic model, then the simplification due to the symmetries and geometries of the robot are carried out. After that, by means of statistical considerations the model is reduced until the physical feasibility conditions are satisfied. The implementation of the proposed methodological strategy were tested over two types of parallel robots with 3-DOF. The strategy yielded a model which has been verified by solving the inverse dynamic problem and subsequent comparison of the generalized forces. Moreover, the formulation of the forward dynamic problem has been dealt with by considering the identified dynamic parameters. In both cases, the system response, based on the subset of relevant parameters, has shown agreement with the actual performances. In addition, the response of the reduced model is more closely to the experiments compared to the full model.

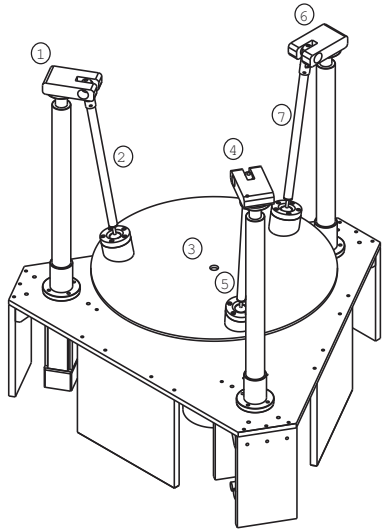

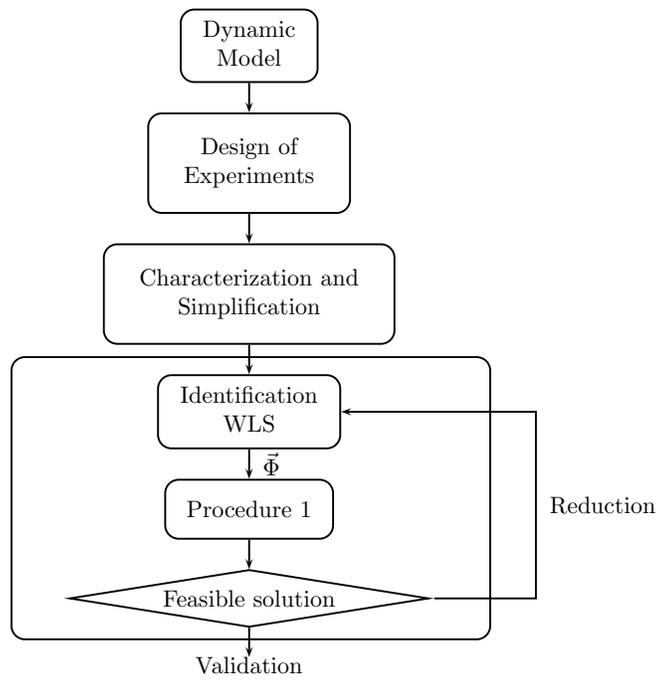

Figure 3: Flowchart of the proposed methodology

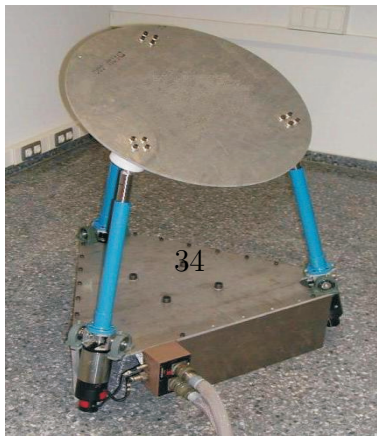

Figure 4: Actual 3-PRS and 3-RPS parallel robots used in the experiments

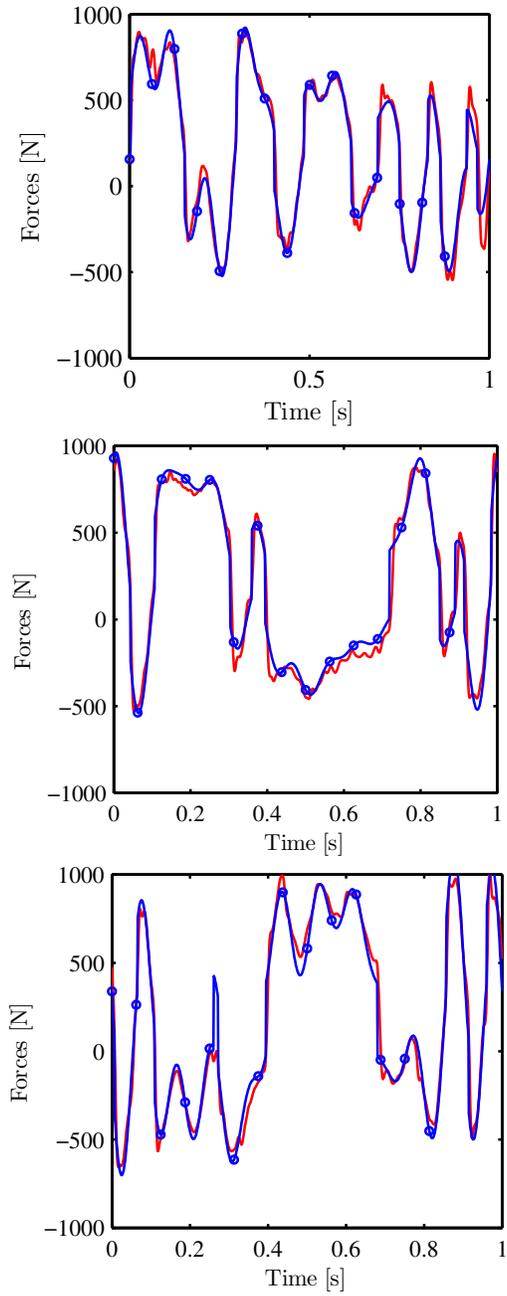

Figure 5: Measured forces (—) and estimated forces (o—) for the 3-PRS robot.

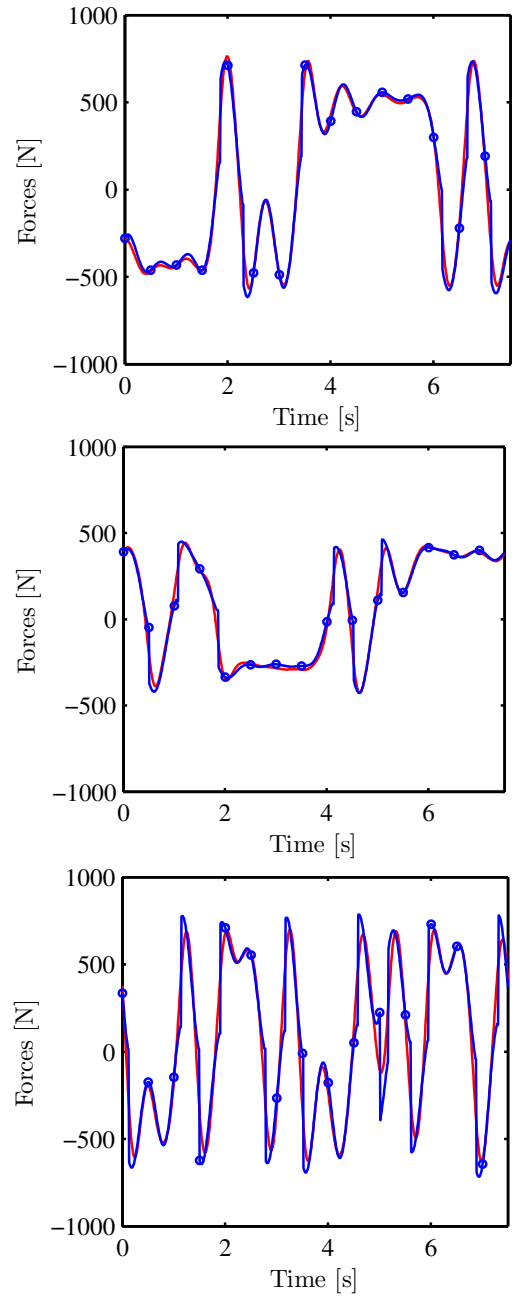

Figure 6: Measured forces (—) and estimated forces (o—) for the 3-RPS robot.

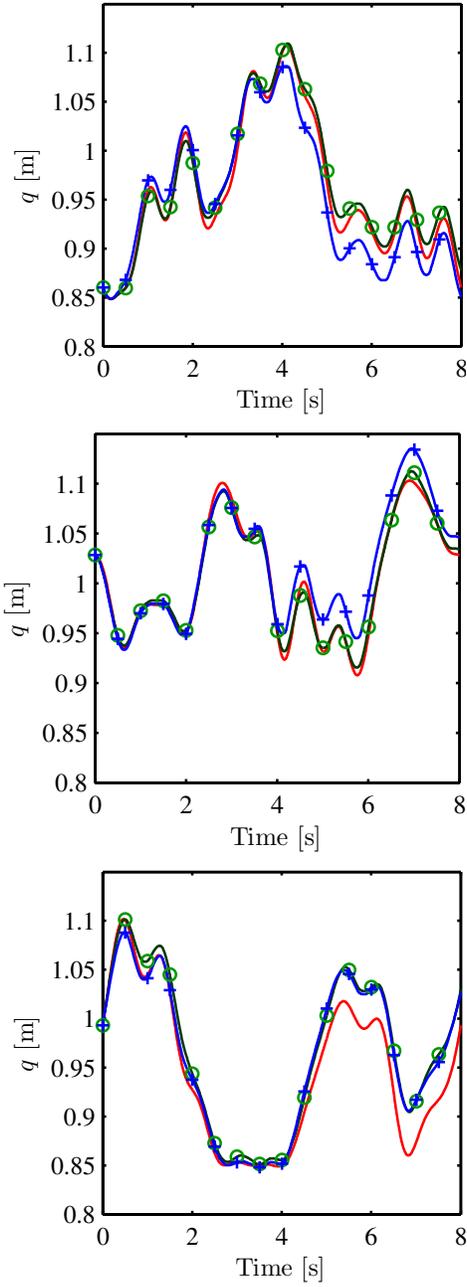

Figure 7: Actual displacement(—), simulated reduced model (o—) and simulated complete model (+—), for the 3-PRS robot.

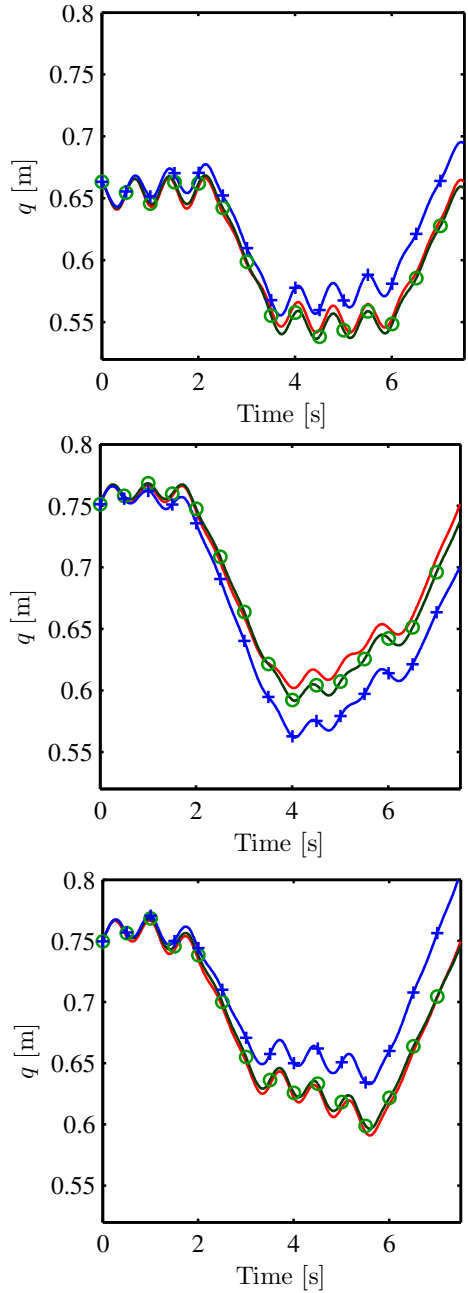

Figure 8: Actual displacement(—), simulated reduced model (o—) and simulated complete model (+—), for the 3-RPS Robot.